\documentclass[a4paper,fleqn]{cas-dc}

\usepackage[numbers]{natbib}

\def\tsc#1{\csdef{#1}{\textsc{\lowercase{#1}}\xspace}}
\tsc{WGM}
\tsc{QE}
\tsc{EP}
\tsc{PMS}
\tsc{BEC}
\tsc{DE}

\usepackage{makecell}
\usepackage[ruled,vlined]{algorithm2e}
\usepackage{hyperref}
\usepackage{cleveref}
\newcommand{\KwStep}{}
\begin{document}
\let\WriteBookmarks\relax
\def\floatpagepagefraction{1}
\def\textpagefraction{.001}

\shorttitle{Explainable Multi-Task Classification in Autonomous Vehicles}

\shortauthors{MSH Azad et~al.}

\title [mode = title]{Beyond Fixed Thresholds and Domain-Specific Benchmarks for Explainable Multi-Task Classification in Autonomous Vehicles}                      


%
\author[1]{Maryam Sadat Hosseini Azad}[
                        orcid=0009-0009-2473-4934]

\cormark[1]


\ead{hosseini_maryam@elec.iust.ac.ir}



\affiliation[1]{organization={Iran University of Science and Technology},
    addressline={School of Electrical Engineering}, 
    city={Tehran},
    country={Iran}}

\author[1]{Shahriar Baradaran Shokouhi}
[orcid=0000-0001-6266-6607]
\ead{bshokouhi@iust.ac.ir}

\cortext[cor1]{Corresponding author}


\begin{abstract}
Scene understanding is a vital part of autonomous driving systems, which requires the use of deep learning models. Deep learning methods are intrinsically black box models, which lack transparency and safety in autonomous driving. To make these systems transparent, multi-task visual understanding has become crucial for explainable autonomous driving perception systems, where simultaneous prediction of multiple driving behaviors and their underlying explanations is essential for safe navigation and human trust in autonomous vehicles. In order to design an accurate and cross-cultural explainable autonomous driving system, we introduce a comprehensive confidence threshold sensitivity analysis that evaluates various threshold values to identify optimal decision boundaries for different tasks. Our analysis demonstrates that traditional fixed threshold approaches are suboptimal for multi-task scenarios. Through extensive evaluation, we demonstrate that our adaptive threshold selection methodology improves F1-scores across different tasks. In addition, we introduce IUST-XAI-AD, a novel dataset consisting of 958 images with human annotations for driving decisions and corresponding reasoning. This dataset addresses the critical gap in domain-specific evaluation benchmarks for distinct driving contexts and provides a more challenging test environment compared to existing datasets. Experimental results demonstrate that confidence threshold sensitivity analysis can significantly improve model performance, while the introduction of the IUST-XAI-AD dataset reveals important insights about cross-cultural driving behavior patterns. The combined contributions of this work provide both methodological advances and practical evaluation tools that can accelerate the development of more reliable, explainable, and culturally-adaptive autonomous driving systems for global deployment.
\end{abstract}




\maketitle

\section{Introduction}
The complexity of deep learning methods that are employed in autonomous driving technology has raised major concerns about their transparency and safety. In order to interpret these deep learning methods, multi-task learning has gained remarkable attention in the development of intelligent systems capable of making complex driving decisions while providing human-understandable explanations for their actions \cite{dong2021image, sun2025semantic}. Modern autonomous vehicles must not only perceive their environment and decide on appropriate actions but also justify these decisions in a manner that builds trust with human operators and regulatory authorities \cite{xie2025human, gyevnar2022human, hosseiniazad2026endtoend}. This dual requirement has led to the emergence of explainable artificial intelligence (XAI) approaches in autonomous driving, where systems simultaneously predict driving actions and provide reasoning for their decisions.

\begin{figure}
    \centering
    \includegraphics[width=1\linewidth]{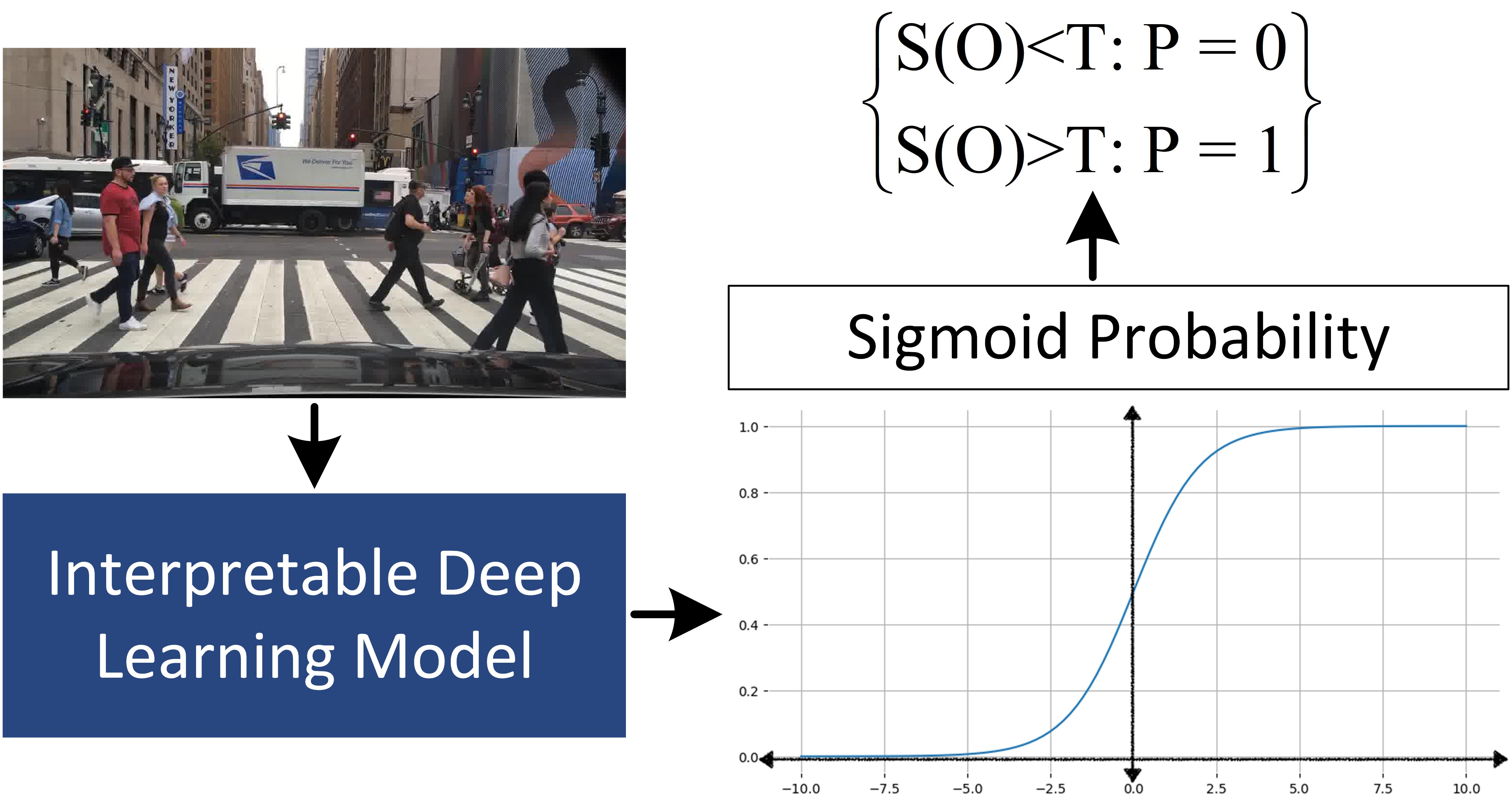}
    \caption{The overview of the studied problem. S refers to sigmoid, O refers to output, T refers to threshold, and P refers to prediction. The workflow processes a dash-cam image through an interpretable deep learning model. The model outputs a sigmoid probability score S(O) that is compared to a threshold T, producing a binary classification prediction P.}
    \label{fig:placeholder1}
\end{figure}

Multi-task learning frameworks have shown remarkable promise in addressing the challenge of joint action-reason prediction in autonomous driving scenarios \cite{cao2024sgdcl, meng2025foundation, fu2024top}. By learning shared representations across related tasks, these models can leverage common visual features while maintaining task-specific decision pathways. However, current approaches in this domain face several critical limitations that hinder their practical deployment and evaluation. First, the majority of existing multi-task classification systems rely on fixed confidence thresholds \cite{wang2019deep,jing2022inaction}, typically set to 0.5, for converting continuous model outputs into discrete predictions. This approach ignores the inherent differences in task complexity and class imbalance across different prediction tasks. In autonomous driving scenarios, where safety-critical decisions must be made with appropriate confidence, this one-size-fits-all approach can lead to suboptimal performance and potentially dangerous false positives or negatives. We hypothesize that optimizing confidence thresholds for action and reason prediction tasks will significantly improve overall system performance compared to conventional fixed-threshold approaches. Second, the evaluation of multi-task models for autonomous driving has been primarily constrained to Western driving contexts and datasets, such as the Berkeley Deep Drive (BDD) \cite{yu2020bdd100k} dataset. While these datasets provide valuable benchmarks, they may not capture the full diversity of driving behaviors, traffic patterns, and cultural variations present in different regions of the world. This limitation raises questions about the generalizability of current approaches and their applicability to global deployment scenarios. We investigate whether models evaluated on culturally diverse datasets demonstrate different performance characteristics and whether cross-cultural evaluation reveals systematic gaps in model robustness that single-domain evaluation might miss. We examine whether a Middle Eastern dataset—featuring different traffic patterns, road infrastructure, and cultural driving behaviors—exposes unique challenges and evaluation requirements that are not adequately captured by existing benchmarks such as Berkeley Deep Drive Object Induced Action (BDD-OIA) \cite{xu2020explainable} and nuScense Action and Reasons (nu-AR) \cite{feng2023nle}. Third, there exists a significant gap in comprehensive validation methodologies that can effectively assess both the technical performance and practical applicability of multi-task learning approaches across different evaluation contexts. Most current studies focus on single-dataset validation, limiting insights into model robustness and cross-domain transferability. The overview of the problem investigated in this study is shown in Figure \ref{fig:placeholder1}. To address these limitations, this paper presents three interconnected contributions that advance the state-of-the-art in explainable multi-task learning for autonomous driving:
\begin{itemize} \item We introduce a systematic approach for optimizing confidence thresholds in multi-task classification systems. Unlike traditional fixed-threshold methods, our framework evaluates multiple thresholds across different tasks to identify optimal decision boundaries. This approach recognizes that different tasks may require different levels of confidence for reliable predictions, particularly in safety-critical applications where false positives and false negatives carry different risk profiles. \item We present a novel dataset specifically designed for evaluating explainable computer vision in autonomous driving. The IUST-XAI-AD dataset comprises 958 carefully annotated images representing diverse driving scenarios encountered in Persian driving contexts. Each image is annotated with detailed action labels and corresponding explanatory reasons, providing a comprehensive benchmark for multi-task learning evaluation. The dataset addresses the cultural and regional bias present in existing benchmarks while maintaining compatibility with established evaluation protocols. \item We conduct extensive complexity evaluations of our dataset across multiple evaluation contexts. By testing our previous model on both the new IUST-XAI-AD dataset and the established BDD-OIA and nu-AR benchmarks, we provide insights into the effectiveness of our dataset.
\end{itemize}

The remainder of this paper is organized as follows. Section \ref{sec:section2} reviews related work. Section \ref{sec:section3} details our confidence threshold sensitivity analysis and the construction of the IUST-XAI-AD dataset. Section \ref{sec:section4} presents our experimental methodology and evaluation protocols. Section \ref{sec:section5} discusses the experimental results, including comparative analysis across different datasets and threshold configurations. Finally, Section \ref{sec:section6} concludes the paper and outlines directions for future research, respectively.

Through this comprehensive investigation, we aim to provide both theoretical insights and practical tools that can accelerate the development of more reliable, explainable, and globally applicable autonomous driving systems. Our work contributes to the broader goal of creating AI systems that not only perform well statistically but also operate safely and transparently in diverse real-world contexts.

\section{Related Work}
\label{sec:section2}
This section reviews the relevant literature across three key areas that form the foundation of our research: multi-task learning in explainable autonomous driving, confidence threshold optimization techniques, and examining existing datasets and evaluation methodologies to position our contributions within the current landscape.

\subsection{Multi-Task Learning for Explainable Autonomous Driving}
Multi-task learning has gained increasingly popular in autonomous driving applications due to its ability to leverage shared visual representations across related prediction tasks \cite{wang2025interpretable, chowdhuri2019multinet}. Early work by Xu et al. \cite{xu2020explainable} demonstrated that joint learning of driving actions and explanations could improve the task performance of decision making compared to independent single-task models. It established a benchmark for evaluating multi-task approaches in real-world driving scenarios. Furthermore, the NLE-DM \cite{feng2023nle} model incorporates fine-grained segmentations to provide more interpretable explanations for driving decisions. However, these approaches typically rely on fixed confidence thresholds for prediction, which may not be optimal for the diverse task requirements in autonomous driving scenarios.

\subsection{Confidence Threshold Optimization and Analysis}
Confidence threshold optimization has been extensively studied in single-task classification problems \cite{bukowski2021decision,leo2021incremental,taha2022confidence,lv2024high,thomas2024ltrc}, but its application to multi-task scenarios remains limited. While these approaches can enhance individual task performance, they do not address the challenge of optimizing thresholds across multiple interdependent tasks simultaneously in explainable autonomous driving. This finding has particular relevance for safety-critical applications, where appropriate confidence threshold determination is crucial for reliable decision-making.

In medical applications, Tang et al. \cite{tang2019interpretable} developed class-specific confidence thresholds for Alzheimer's disease classification (0.91 for cored plaques, 0.1 for diffuse plaques, and 0.85 for Cerebral Amyloid Angiopathy), achieving 0.993 AUROC (Area Under the Receiver Operating Characteristic) for cored plaque detection. Wada et al. \cite{wada2024optimizing} demonstrated that implementing 90\% confidence thresholds in GPT-4 Turbo for neuroradiology diagnosis improved precision from 55.1\% to 72.9\%, though only 47\% of cases met the high-confidence criteria. Zhang et al. \cite{zhang2020viral} proposed confidence-aware anomaly detection for viral pneumonia screening, improving sensitivity from 85.51\% to 93.01\%.

On the other hand, recent advances have focused on dynamic threshold adaptation. Thatikonda et al. \cite{thatikonda2024novel} introduced Dynamic Confidence Threshold (DCT) algorithms that adjust thresholds based on image statistical properties rather than fixed values, achieving 56.11\% mAP on the COCO dataset with Faster R-CNN \cite{cheng2018revisiting}. Ma et al. \cite{van2023adaptive} addressed ByteTrack's fixed threshold limitations by proposing adaptive methods using steepest descent analysis of confidence score distributions. EdgeBERT \cite{tambe2021edgebert} employed entropy-based confidence thresholds for early exit prediction in BERT \cite{devlin2019bert} inference, enabling dynamic voltage-frequency scaling for energy-efficient natural language processing on edge devices.

Specialized applications have developed domain-specific thresholding strategies. Qi et al. \cite{qi2018adaptive} leveraged temporal correlations in video for adaptive vehicle detection thresholds. Gomez et al. \cite{gomez2016animal} implemented "alleviation mode" for animal species classification, delegating low-confidence predictions to human experts, achieving 100\% accuracy through human-AI collaboration. Industrial applications have also benefited from confidence thresholds. Bassani et al. \cite{bassani2025leveraging} established Predicted Goodness thresholds of 9-10 for ADME-Tox pharmaceutical assays, achieving 9-25\% reduction in experimental testing while maintaining accuracy within acceptable bounds. The summary of previous research is shown in Table \ref{tbl1}.

\begin{table*}[t]
\caption{Summary of Previous Research on Confidence Threshold Optimization and Analysis.}\label{tbl1}
\centering
\begin{tabular*}{\textwidth}{@{\extracolsep{\fill}} l p{5.95cm} c p{5.9cm} @{}}
\toprule
Work & Dataset & \makecell{Number of \\ Classes} & Thresholding Strategy\\
\midrule
Tang et al. \cite{tang2019interpretable} & Whole Slide Images (WSIs) of temporal gyri & 3 & Different threshold numbers for each class \\
Wada et al. \cite{wada2024optimizing} & 751 publicly available neuroradiological cases & 10 & Confidence-Based Thresholding Strategy \\
Zhang et al. \cite{zhang2020viral} & X-VIRAL dataset & 2 & Dual-Threshold Approach \\
Thatikonda et al. \cite{thatikonda2024novel} & COCO dataset & 80 & Dynamic Thresholding Approach \\
Ma et al. \cite{van2023adaptive} & MOT16, MOT17, MOT18 & 1 & Adaptive Confidence Threshold \\
EdgeBERT \cite{tambe2021edgebert} & GLUE benchmark & 2 and 3 & Entropy-Based Thresholding Strategy \\
Qi et al. \cite{qi2018adaptive} & XJTU dataset, ChangSha dataset & 1 & Adaptive Confidence Threshold \\
Gomez et al. \cite{gomez2016animal} & Colombian Camera-Trap dataset & 2 & Alleviation Mode \\
Bassani et al. \cite{bassani2025leveraging} & vitro ADME-Tox datasets & 2 & Based on Predicted Goodness Strategy \\
\bottomrule
\end{tabular*}
\end{table*}

Despite these advances, significant gaps remain in the multi-task classification confidence threshold in explainable autonomous driving contexts. Most existing approaches focus on independent label prediction rather than structured multi-task scenarios where tasks are interdependent and share common representations. The challenge of optimizing thresholds across multiple interdependent tasks simultaneously, particularly in safety-critical applications like autonomous driving, remains largely unaddressed. This gap motivates our comprehensive analysis of confidence thresholds in multi-task autonomous driving scenarios.

\subsection{Datasets and Evaluation Methodologies}
The Berkeley Deep Drive dataset \cite{yu2020bdd100k} has served as a primary benchmark for autonomous driving research, providing large-scale annotations for various driving tasks, including object detection, semantic segmentation, and trajectory prediction. The BDD-OIA \cite{xu2020explainable} extension specifically addresses action-explanation prediction by providing structured annotations for driving actions and their corresponding reasons. The BDD-X dataset \cite{kim2018textual} includes generative natural language explanations for driving actions, enabling the development of models that can generate human-readable explanations for autonomous driving decisions. The A2D2 (Audi Autonomous Driving Dataset) \cite{geyer2020a2d2} provides a comprehensive multimodal dataset featuring synchronized data from six cameras and five LiDAR sensors with full 360-degree coverage, along with semantic segmentation labels, 3D bounding boxes, and vehicle bus data. Gadd et al. \cite{gadd2020sense} introduced the Sense-Assess-eXplain (SAX) paradigm for building trust in autonomous vehicles through alternative sensing modalities, continuous performance assessment, and human-interpretable causal explanations of vehicle decisions in challenging weather and terrain conditions. The Waymo Open Dataset \cite{sun2020scalability} is a large-scale multimodal autonomous driving dataset containing 1,150 scenes with synchronized LiDAR and camera data across diverse geographical areas, providing 12M 3D LiDAR annotations and 9.9M 2D camera annotations for object detection and tracking tasks. The Honda Driving Dataset (HDD) \cite{ramanishka2018toward} is a comprehensive multimodal dataset containing 104 hours of real human driving data collected in the San Francisco Bay Area using an instrumented vehicle equipped with cameras, LiDAR, GPS, IMU, and CAN sensors. The PSI (Pedestrian-Scene Interaction) dataset \cite{chen2021psi} provides novel annotations for dynamic pedestrian intent changes during vehicle-pedestrian interactions, along with textual explanations of driver reasoning processes. The Cityscapes dataset \cite{cordts2016cityscapes} is a large-scale benchmark containing 5,000 finely annotated and 20,000 coarsely annotated urban street scene images, specifically designed for semantic segmentation and instance-level object detection in autonomous driving scenarios. The KITTI dataset \cite{geiger2013vision} is a comprehensive autonomous driving dataset captured from a Volkswagen station wagon equipped with stereo cameras, a Velodyne 3D laser scanner, and GPS/IMU sensors, containing 6 hours of traffic scenarios with calibrated, synchronized data and 3D object annotations. The nuScenes dataset \cite{caesar2020nuscenes} is a large-scale autonomous driving dataset that provides a complete 360-degree sensor suite, including 6 cameras, 5 radars, and 1 lidar, comprising 1000 scenes with 1.4M 3D bounding box annotations across 23 object classes.

Particularly notable is the absence of datasets representing Middle Eastern driving contexts for explainable autonomous driving. This gap is significant given the unique characteristics of distinct driving environments, including different traffic patterns, road infrastructure design, and cultural driving behaviors that may not be well-represented in existing Western-centric datasets.

Our work addresses these gaps through the introduction of systematic confidence threshold sensitivity analysis for multi-task scenarios, the development of a culturally diverse evaluation dataset, and comprehensive validation across multiple benchmark contexts. These contributions provide both methodological advances and practical tools that can accelerate the development of more reliable and globally applicable autonomous driving systems.

\section{Model Architecture and Methodology}
\label{sec:section3}
This section presents the overall architecture, our novel confidence threshold sensitivity analysis, and the IUST-XAI-AD dataset construction. We build upon our previously proposed attention-based approach \cite{azad2024xai} while introducing systematic methods for optimizing decision boundaries across multiple related tasks.

\subsection{Overall Architecture}
Our multi-task model follows a CNN-based architecture with shared feature extraction and task-specific prediction heads. The overall framework consists of four main components: (1) a ResNet-50 backbone for feature extraction, (2) an attention-enhanced feature processing module, (3) task-specific classification heads for action and reason prediction, and (4) a confidence threshold analysis module. The core architecture builds upon our previously proposed attention-based multi-task learning model \cite{azad2024xai}, which demonstrated effective performance for joint action-reason prediction. The base model employs a ResNet-50 backbone enhanced with attention mechanisms to learn discriminative features for simultaneous action and reason prediction. The attention-based feature processing module applies channel and spatial attention mechanisms to focus on relevant image regions for decision-making. The shared feature extractor learns common visual representations, while task-specific heads ensure that action and reason predictions maintain their distinct characteristics while benefiting from shared learning. Rather than modifying the proven architecture, our key contribution lies in the systematic analysis of confidence thresholds and comprehensive evaluation on culturally diverse datasets.

\subsection{Confidence Threshold Sensitivity Analysis Framework}
A key contribution of this work is the systematic analysis of confidence thresholds for multi-task classification. Traditional approaches use fixed thresholds (typically 0.5) for converting continuous outputs to binary predictions. However, this approach may be suboptimal for multi-task scenarios where different tasks have varying difficulty levels and class distributions.
For a multi-task classification system with action prediction task A and reason prediction task R, let:
\setlength{\mathindent}{0pt}
\begin{equation*}
P_A(X)\in[0,1]^{\tau_A} \text{represent the action prediction probabilities}
\end{equation*}
\vspace{-20pt}
\setlength{\mathindent}{0pt}
\begin{equation*}
P_R(X)\in[0,1]^{\tau_R} \text{represent the reason prediction probabilities}
\end{equation*}
$\tau_A$ and $\tau_R$ denote the confidence thresholds for action and reason tasks, respectively.
We define the threshold search space as:
\setlength{\mathindent}{0pt}
\begin{equation*}
\text{Action thresholds:} \tau_A \in [0,1] \text{ with step size 0.1} 
\end{equation*}
\vspace{-20pt}
\setlength{\mathindent}{0pt}
\begin{equation*}
\text{Reason thresholds:} \tau_R \in [0,1] \text{ with step size 0.1}
\end{equation*}
Our framework optimizes task-specific thresholds:
\setlength{\mathindent}{0pt}
\begin{equation}
\hat{y}_A  =1; \ if \ P_A (X)>\tau_A, else \ 0
\end{equation}
\vspace{-20pt}
\setlength{\mathindent}{0pt}
\begin{equation}
\hat{y}_R  =1; \ if \ P_R (X)>\tau_R, else \ 0
\end{equation}
For each threshold $\tau_A$ and $\tau_R$, we evaluate model performance using multiple metrics:
(1) F1-action-overall: Sample-wise F1 score across action samples
(2) F1-action-mean: Class-wise F1 score across action classes
(3) F1-reason-overall: Sample-wise F1 score across reason samples
(4) F1-reason-mean: Class-wise F1 score across reason samples

Algorithm \ref{alg:threshold} systematically searches through threshold combinations for classes A and R to find the optimal pair that maximizes the F1 scores. It uses a nested loop approach with a configurable step size to evaluate threshold combinations within the [0,1] range, tracking the best-performing configuration based on F1 scores across both classes. This approach allows practitioners to select thresholds based on their specific operational requirements rather than forcing a predetermined trade-off through a weighted combination.
We selected $\delta = 0.1$ based on three considerations:
\begin{enumerate}
\itemsep=0pt
    \item Computational efficiency: 36 evaluations versus 72 for $\delta=0.05$
    \item Practical significance: Our analysis (Table \ref{tbl3}) shows performance changes smoothly, with differences between adjacent grid points < 1\% in the robust region
    \item For applications requiring finer-grained optimization, our results indicate the search should focus on [0.3, 0.5].
\end{enumerate}

\begin{algorithm}[t]
\caption{Confidence Threshold Sensitivity Analysis}
\label{alg:threshold}
\KwIn{Model $M$, Validation set $V$, Threshold range $[0, 1]$, Step size $\delta$}
\KwOut{Performance landscape for all $(\tau_A, \tau_R)$ combinations}

Initialize results matrix $R[9 \times 9]$ for metrics\;

\For{$\tau_A = 0$ \KwTo $1$ \KwStep $\delta$}{
    \For{$\tau_R = 0$ \KwTo $1$ \KwStep $\delta$}{
        Apply thresholds $(\tau_A, \tau_R)$ to model predictions\;
        
        Compute and store\;
        \Indp
        F1-action-overall, F1-action-mean\;
        F1-reason-overall, F1-reason-mean\;
        \Indm
    }
}

\Return performance landscape $R$\;

\BlankLine
\textbf{Post-analysis:}

(1) Identify peak performance for each metric

(2) Characterize robust threshold range

(3) Visualize trade-offs between tasks
\end{algorithm}

This systematic approach identifies appropriate operating points for each task while considering their interdependencies. While our primary contribution focuses on systematic threshold analysis, we explore non-trainable adaptive mechanisms that can adjust confidence thresholds based on system state and environmental conditions without requiring additional model training.

\subsection{IUST-XAI-AD Dataset Construction}
\begin{figure}
    \centering
    \includegraphics[width=1\linewidth]{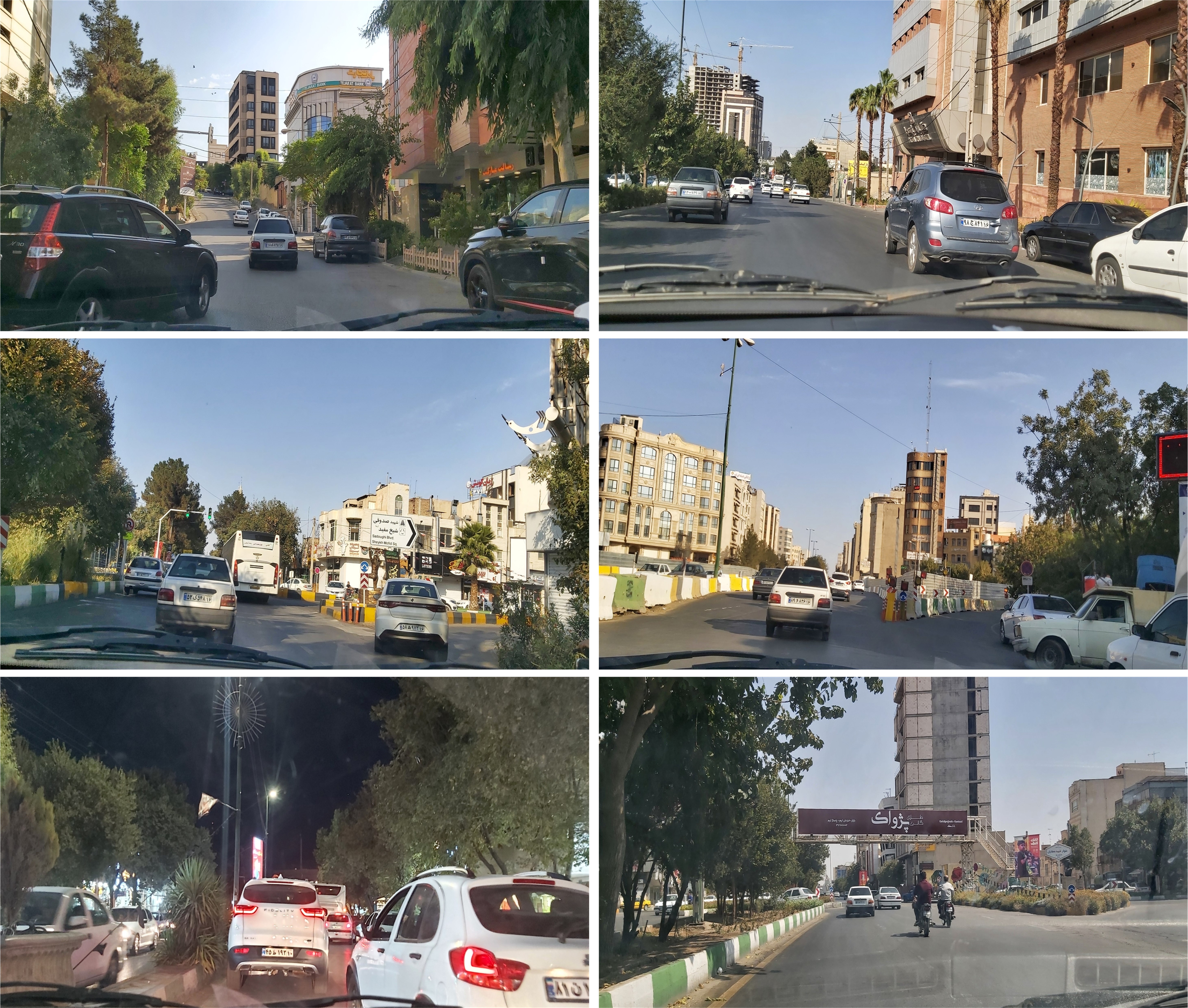}
    \caption{Several frames from the IUST-XAI-AD dataset captured at different locations and times of day, demonstrating the high complexity and diversity of the dataset.}
    \label{fig:placeholder2}
\end{figure}

The second major contribution of this work is the introduction of the IUST-XAI-AD dataset for comprehensive model validation across different cultural contexts. The IUST-XAI-AD dataset was collected during summer 2025 in Qom city, Iran, capturing images at various times throughout the day and night under different lighting conditions from complex driving scenarios. The dataset comprises 958 images with comprehensive human annotations for both action and reason classification tasks. It consists of 4 categories of action and 21 categories of reason. All labels were manually annotated for each image by expert human annotators, ensuring high-quality ground truth annotations for both action and reasoning tasks. Table \ref{tbl2} shows the distribution comparison across three test datasets. In addition, several frames from the IUST-XAI-AD dataset are demonstrated in Figure \ref{fig:placeholder2}. The class distribution analysis reveals the higher frequency of "Stop/slow down" actions in Persian driving contexts (56.1\% versus 46.0\% in BDD-OIA). Therefore, Cultural driving differences are reflected in action distribution patterns.

\begin{table*}[width=2.05\linewidth,cols=8,pos=h]
\caption{Class Distribution Comparison Across Test Datasets. Comparison of action and reason class distributions across BDD-OIA \cite{xu2020explainable}, nu-AR \cite{feng2023nle}, and IUST-XAI-AD test datasets. The table shows the frequency and percentage of each action and reason category.}\label{tbl2}
\begin{tabular*}{\tblwidth}{@{} L C C C L C C C @{} }
\toprule
Action Category & \makecell{BDD-OIA \\ (\%)} & \makecell{Nu-AR \\ (\%)} & \makecell{IUST-XAI-AD \\ (\%)} & Reason Category & \makecell{BDD-OIA \\ (\%)} & \makecell{Nu-AR \\ (\%)} & \makecell{IUST-XAI-AD \\ (\%)} \\
\midrule
Move forward & 54.3 & 71.6 & 41.3 & \makecell[l]{Follow traffic \\ Road is clear \\ Traffic light is green} & \makecell{34.4 \\ 14.6 \\ 20.6} & \makecell{23.4 \\ 49.5 \\ 17.8} & \makecell{30.4 \\ 2.6 \\ 1.6} \\
\addlinespace
Stop/Slow down & 46.0 & 28.4 & 56.1 & \makecell[l]{Obstacle: car \\ Obstacle: person/pedestrian \\ Obstacle: rider \\ Obstacle: others \\ Traffic light \\ Traffic sign} & \makecell{23.7 \\ 6.6 \\ 0.9 \\ 0.74 \\ 22.7 \\ 2} & \makecell{10.1 \\ 7.2 \\ 0.7 \\ 0.4 \\ 23.0 \\ 0.1} & \makecell{41.0 \\ 8.9 \\ 16.4 \\ 0.2 \\ 6.4 \\ 0.0} \\
\addlinespace
Turn left & 26.8 & 32.4 & 14.4 & \makecell[l]{Front car turning left \\ On the left-turn lane \\ Traffic light allows} & \makecell{0.6 \\ 3 \\ 1.5} & \makecell{0.3 \\ 0.2 \\ 0.0} & \makecell{0.3 \\ 1.9 \\ 0.0} \\
\addlinespace
Turn right & 29.3 & 33.1 & 5.9 & \makecell[l]{Front car turning right \\ On the right-turn lane \\ Traffic light allows} & \makecell{0.5 \\ 3.8 \\ 1.5} & \makecell{0.0 \\ 0.8 \\ 0.3} & \makecell{0.4 \\ 1.1 \\ 0.0} \\
\addlinespace
\makecell[l]{Can't change \\ to left lane} & - & - & - & \makecell[l]{Obstacles on the left lane \\ No lane on the left \\ Solid line on the left} & \makecell{18.9 \\ 19.7 \\ 15.6} & \makecell{14.7 \\ 31.4 \\ 32.3} & \makecell{53.1 \\ 45.1 \\ 12.0} \\
\addlinespace
\makecell[l]{Can't change \\ to right lane} & - & - & - & \makecell[l]{Obstacles on the right lane \\ No lane on the right \\ Solid line on the right} & \makecell{25.9 \\ 18.1 \\ 8.9} & \makecell{20.8 \\ 23.8 \\ 31.8} & \makecell{70.9 \\ 26.2 \\ 8.1} \\
\bottomrule
\end{tabular*}
\end{table*}

To quantitatively assess the complexity of different datasets, we define a weighted complexity score that accounts for the varying difficulty of detecting and reasoning about different object types in autonomous driving scenarios. The complexity score C is calculated as:
\setlength{\mathindent}{0pt}
\begin{equation}
C = 1.5\times D_p+1.3\times D_r+1.0\times D_v
\end{equation}
Where $D_p$, $D_r$, and $D_v$ represent the density (objects per image) of pedestrians, riders, and vehicles, respectively. The weighting scheme reflects the relative challenge each object type poses for autonomous driving systems: pedestrians (weight = 1.5) require the highest attention due to their unpredictable movement patterns and vulnerability; riders (weight = 1.3) present intermediate complexity due to their smaller profile and faster movement compared to vehicles; and vehicles (weight = 1.0) serve as the baseline complexity unit due to their larger size, more predictable trajectories, and established detection methods. This weighting is consistent with established autonomous driving safety frameworks, where vulnerable road users (VRUs) such as pedestrians and riders are assigned higher priority in risk assessment and decision-making protocols \cite{yuan2023summarizing, muslim2022review, he2025configurable}.

\section{Evaluation Metrics}
\label{sec:section4}
To comprehensively evaluate our multi-task learning framework and confidence threshold analysis, we employ multiple metrics that capture different aspects of model performance and mitigate the challenge of imbalanced datasets.

To compute the F1 score, we first define the fundamental classification metrics. Precision measures the accuracy of positive predictions, representing the proportion of correctly predicted positive instances among all instances predicted as positive:
\setlength{\mathindent}{0pt}
\begin{equation}
Precision =  \frac{TP}{(TP+FP)}
\end{equation}
where TP (True Positives) represents correctly identified positive cases and FP (False Positives) represents negative cases incorrectly classified as positive. High precision indicates that the model makes few false positive errors, which is crucial in autonomous driving to avoid unnecessary interventions.

Recall measures the ability of the model to identify all positive instances, representing the proportion of correctly predicted positive instances among all actual positive instances:
\setlength{\mathindent}{0pt}
\begin{equation}
Recall =  \frac{TP}{(TP+FN)}
\end{equation}
where FN (False Negatives) represents positive cases incorrectly classified as negative. High recall ensures that the model captures most of the critical driving situations, which is essential for safety in autonomous driving scenarios.

The F1 score provides a harmonic mean of precision and recall, offering a single metric that balances both measures. Unlike the arithmetic mean, the harmonic mean penalizes extreme values, ensuring that a good F1 score requires both precision and recall to be reasonably high:
\begin{equation}
F1 = \frac{2\times(Precision\times Recall)}{(Precision+Recall)}=\frac{2TP}{(2TP+FP+FN)}
\end{equation}
\setlength{\mathindent}{0pt}
This metric is particularly valuable when dealing with imbalanced datasets or when both false positives and false negatives carry significant costs, as is the case in driving action prediction.

Mean and overall F1 score for action are evaluated separately for the four driving actions (move forward, stop/slow down, turn left, turn right), and all samples of the dataset, respectively:
\begin{equation}
Mean \ F1 \ score \ for \ Action=  \frac{1}{|n_A |}\sum_{j=1}^{|n_A |}F1_j^A 
\end{equation}
\vspace{-10pt}
\begin{equation}
\setlength{\mathindent}{0pt}
Overall \ F1 \ score \ for \ Action=  \frac{1}{|N_A |}\sum_{i=1}^{|N_A |}F1(\hat{A}_i,A_i)
\end{equation}
Where $n_A$ is the number of action categories and $N_A$ denotes the number of action predictions. The mean F1 score provides a category-wise evaluation, revealing which specific driving actions the model handles more effectively, while the overall F1 score averages performance across all samples.

Mean and overall F1 score for reason are computed for the 21 explanatory categories, and all samples of the dataset, respectively:
\setlength{\mathindent}{0pt}
\begin{equation}
Mean \ F1 \ score \ for \ Reason=  \frac{1}{|n_R |}\sum_{j=1}^{|n_R |}F1_j^R 
\end{equation}
\vspace{-10pt}
\setlength{\mathindent}{0pt}
\begin{equation}
Overall \ F1 \ score \ for \ Reason=  \frac{1}{|N_R |}\sum_{i=1}^{|N_R |}F1(\hat{R}_i,R_i)
\end{equation}
Where $n_R$ is the number of reason categories and $N_R$ denotes the number of reason predictions. The reasoning task is inherently more challenging than action prediction due to the larger number of categories and the complex nature of explanatory labels, making these metrics particularly important for evaluating the interpretability of the model and trustworthiness in autonomous driving applications.

\section{Experimental Results}
\label{sec:section5}
This section presents comprehensive experimental results demonstrating the effectiveness of our confidence threshold sensitivity analysis and the validation of our multi-task architecture on both established and novel datasets. Our analysis encompasses quantitative performance metrics, comparative evaluations, and detailed insights into the behavior of different threshold configurations. Also, we demonstrate the complexity of our introduced dataset. Our experiments are conducted on three datasets:
\begin{enumerate}
\itemsep=0pt
\item BDD-OIA \cite{xu2020explainable} Dataset: Subset of Berkeley Deep Drive with action-reason annotations consisting of 4,548 images (baseline comparison);
\item nu-AR \cite{feng2023nle} Dataset: nuScense Action and Reasons with 1,502 images (baseline comparison);
\item IUST-XAI-AD Dataset: Our newly introduced dataset with 958 images and human annotations;
\end{enumerate}
All datasets follow the same label structure: 4 action classes (move forward, stop/slow down, turn left, turn right) and 21 reason classes covering various driving scenarios.

\subsection{Threshold Sensitivity Analysis}
Analysis of Table \ref{tbl3} shows that the peak overall action performance occurs at threshold 0.3 (F1-action-overall: 71.85) rather than the conventional 0.5 (71.35). For overall reason prediction, optimal performance is achieved at threshold 0.4 (F1-reason-overall: 54.77) compared to 54.06 at 0.5 performance. Additionally, peak mean action performance occurs at threshold 0.5 (F1-action-mean: 69.59), which equals the conventional 0.5, and the mean reason prediction shows optimal performance at threshold 0.4 (F1-reason-mean: 37.62) compared to 36.65 at 0.5 performance. Degradation is particularly severe at high thresholds (0.8-0.9), with the reduction of 18.67 in the F1-reason-overall, 13.65 in the F1-reason-mean, and 9.23, as well as 4.26 for overall and mean F1 score for action, respectively. This emphasizes the critical importance of proper threshold selection in multi-task classifications. The results reveal significant variations in performance across different threshold values, demonstrating that the conventional 0.5 threshold is suboptimal for multi-task scenarios. The optimal thresholds differ significantly from the conventional 0.5 value. At low thresholds (0.1-0.2), while F1-action metrics remain relatively stable, reasoning metrics suffer significantly. F1-reason-overall drops to 45.37 at threshold 0.1, and F1-reason-mean falls to 32.21. At high thresholds (0.7-0.9), action-related metrics (F1-action-overall, F1-action-mean) are more robust to threshold changes than reason-related metrics. This suggests that action classification is less sensitive to confidence thresholds than reasoning classification.

Our analysis reveals that optimal thresholds vary depending on the prioritized metric:
\begin{itemize}
    \item F1-action-overall peaks at $\tau_A=0.3$ (71.85)
	\item F1-action-mean peaks at $\tau_A=0.5$ (69.59)
	\item F1-reason-overall peaks at $\tau_R=0.4$ (54.77)
	\item F1-reason-mean peaks at $\tau_R=0.4$ (37.62)
\end{itemize}
Importantly, we observe a robust operating region in the range [0.3, 0.5] where all metrics achieve near-optimal performance (within 1\% of their peaks). This finding is practically significant: rather than requiring precise threshold tuning, practitioners can select any threshold in this range based on application-specific priorities without substantial performance degradation.

This robustness contrasts sharply with extreme thresholds (< 0.2 or > 0.7), where at least one metric degrades drastically. The existence of this robust region suggests that the conventional fixed threshold of 0.5 falls within an acceptable range, though our analysis shows task-specific tuning can yield improvements.

\begin{table*}[width=2\linewidth,cols=10,pos=h]
\caption{Threshold Sensitivity Analysis on BDD-OIA \cite{xu2020explainable} Dataset Based on Our Previously Proposed Model \cite{azad2024xai}. Performance metrics (F1-action-overall, F1-action-mean, F1-reason-overall, F1-reason-mean) across different confidence threshold values from 0.1 to 0.9. The best results are specified in \textbf{bold}.}\label{tbl3}
\begin{tabular*}{\tblwidth}{@{} LLLLLLLLLL@{} }
\toprule
Confidence Threshold & 0.1 & 0.2 & 0.3 & 0.4 & 0.5 & 0.6 & 0.7 & 0.8 & 0.9 \\
\midrule
F1-action-overall & 71.25 & 71.78 & \textbf{71.85} & 71.72 & 71.35 & 70.49 & 69.30 & 66.83 & 62.62 \\
F1-action-mean & 68.33 & 69.08 & 69.32 & 69.53 & \textbf{69.59} & 69.25 & 68.77 & 67.56 & 65.33 \\
F1-reason-overall & 45.37 & 51.65 & 54.17 & \textbf{54.77} & 54.06 & 52.03 & 49.26 & 44.55 & 36.10 \\
F1-reason-mean & 32.21 & 35.59 & 37.44 & \textbf{37.62} & 36.65 & 33.93 & 32.41 & 29.18 & 23.97 \\
\bottomrule
\end{tabular*}
\end{table*}

\begin{figure}
    \centering
    \includegraphics[width=1\linewidth]{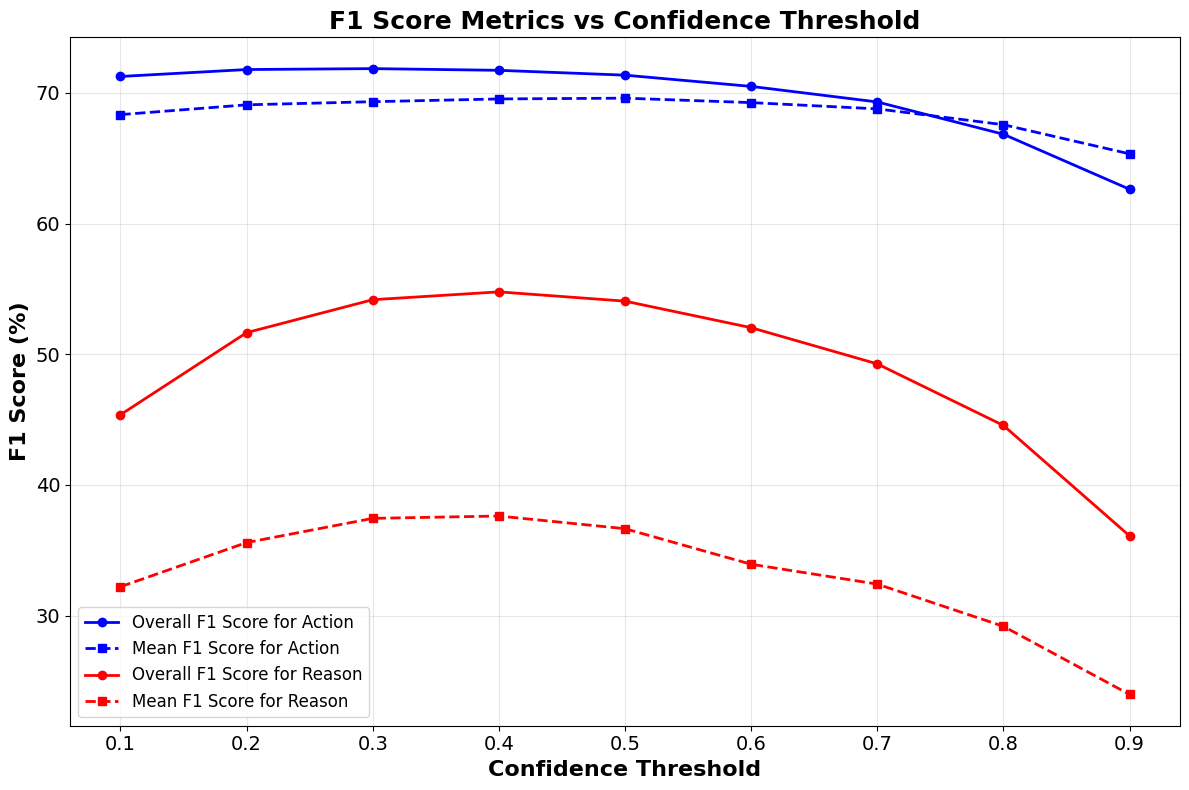}
    \caption{F1 score performance comparison between action (blue lines) and reason (red lines) classification tasks across different confidence thresholds.}
    \label{fig:placeholder3}
\end{figure}

Figure \ref{fig:placeholder3} shows the relationship between confidence thresholds and F1 score performance metrics for two different categories: Action and Reason. The chart displays four metrics: Overall F1 Score for Action (solid blue line with circles), Mean F1 Score for Action (dashed blue line with squares), Overall F1 Score for Reason (solid red line with circles), and Mean F1 Score for Reason (dashed red line with squares). The x-axis represents confidence thresholds ranging from 0.1 to 0.9, while the y-axis shows F1 scores as percentages from approximately 25\% to 75\%. The Action scores remain relatively stable until around 0.7, then decline sharply, while Reason scores peak around 0.3-0.4 confidence threshold before gradually decreasing. This visualization helps identify optimal confidence thresholds for maximizing F1 performance in both categories. Both overall and mean F1 scores for action classification (blue lines) remain relatively stable across most confidence thresholds, maintaining scores around 70\%. However, there's a notable decline at higher thresholds (0.8-0.9), where performance drops to approximately 62-65\%. The reason classification task (red lines) shows more dramatic variation. Performance starts low at 0.1 threshold (~32\% for mean, ~45\% for overall), peaks around 0.3-0.4 thresholds (~37\% for mean, ~54\% for overall), then declines significantly at higher thresholds, dropping to about 24\% (mean) and 36\% (overall) at 0.9. For both tasks, the optimal performance appears to be in the 0.3-0.5 confidence threshold range, where F1 scores are maximized while maintaining reasonable confidence levels.

The relationship between precision and recall when adjusting confidence thresholds represents a fundamental trade-off in machine learning model evaluation:
\begin{itemize}
    \item Increasing Confidence Threshold:
    (1) Precision increases: When raising the threshold, the model only assigns positive labels to predictions with high confidence scores. This approach reduces the number of false positives, as the model becomes more selective in its positive predictions.
    (2) Recall decreases: A higher threshold causes the model to miss many true positive instances that have lower confidence scores, thereby increasing false negatives and reducing the sensitivity of the model.
    \item Decreasing Confidence Threshold:
    (1) Precision decreases: A lower threshold leads the model to classify more instances as positive, including those with lower confidence scores. This increases false positives as the model becomes less discriminating.
    (2) Recall increases: The model captures more true positive instances that were previously missed, reducing false negatives and improving the ability of the model to identify all relevant cases.
    \end{itemize}
    
\begin{figure}
    \centering
    \includegraphics[width=1\linewidth]{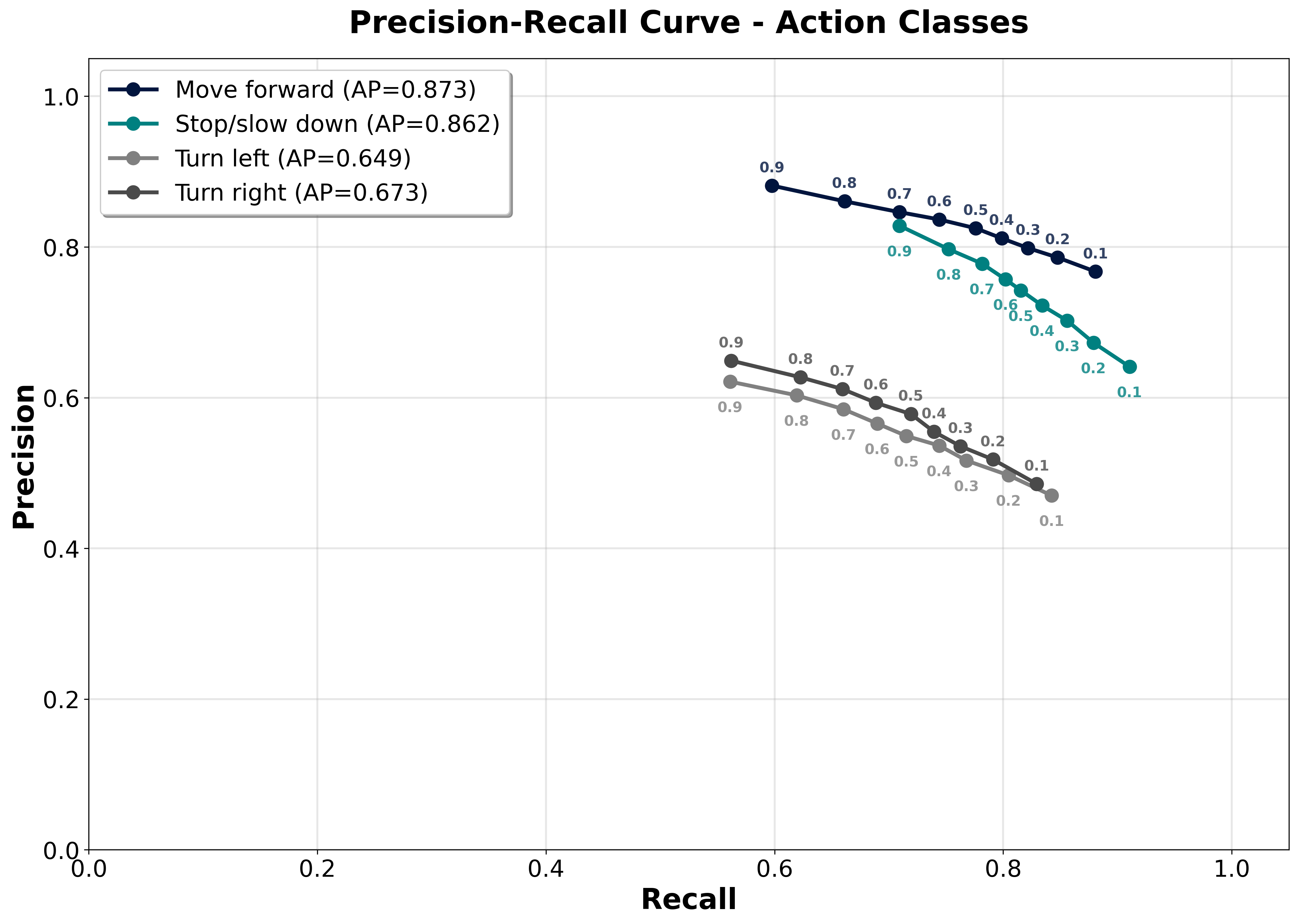}
    \caption{Precision-Recall curves for Action classes. PR curves for four action classes showing high performance with AP scores ranging from 0.649 to 0.873. Points along curves indicate different confidence thresholds (0.1-0.9). "Move forward" and "Stop/slow down" demonstrate superior classification capability.}
    \label{fig:placeholder4}
\end{figure}

\begin{figure}
    \centering
    \includegraphics[width=1\linewidth]{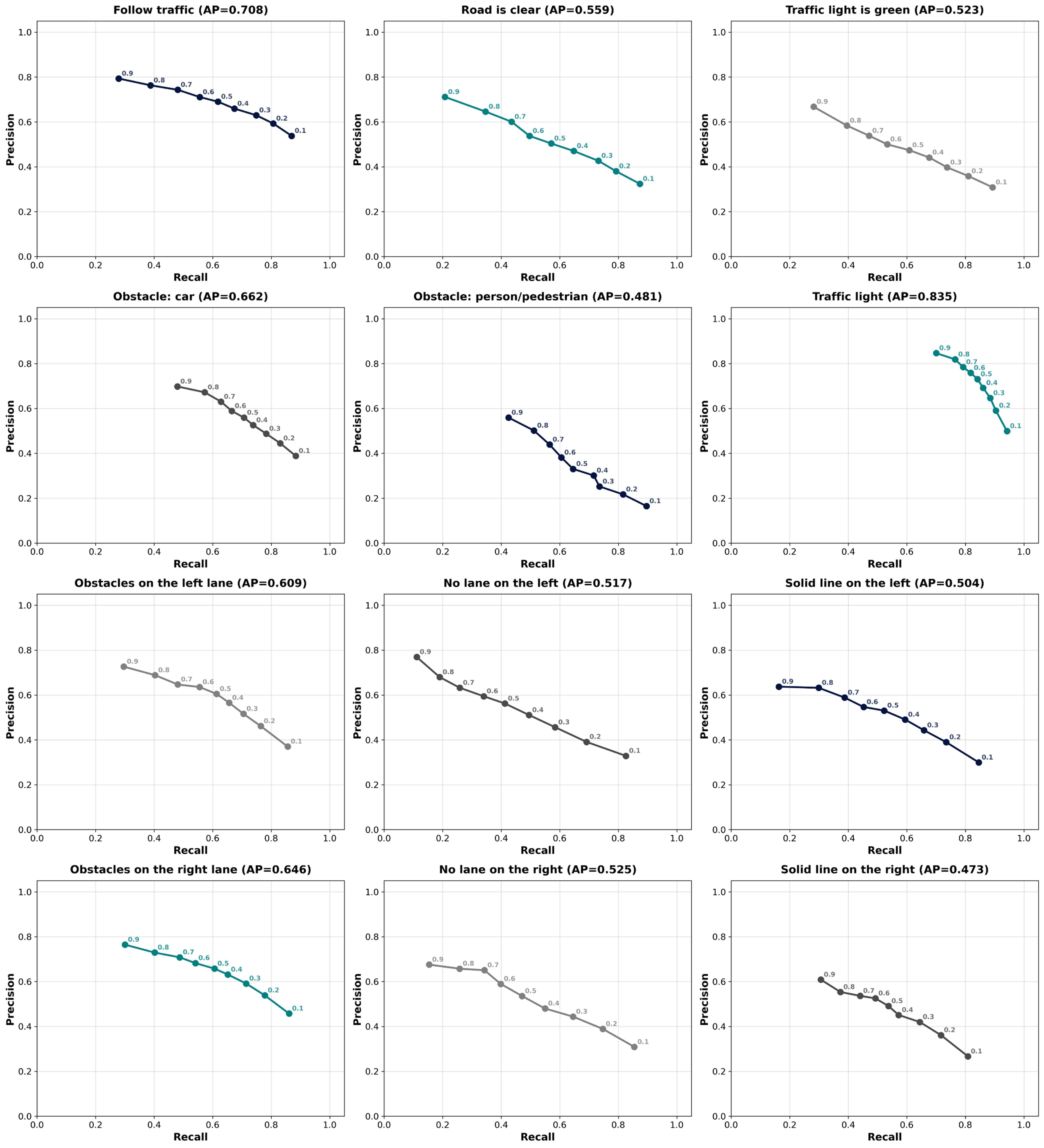}
    \caption{Precision-Recall curves for Reason classes. Each subplot represents a different driving scenario with AP scores in parentheses. Points along curves indicate different confidence thresholds (0.1-0.9). The model achieves the highest performance on traffic following (AP=0.708) and traffic light detection (AP=0.835).}
    \label{fig:placeholder5}
\end{figure}

This inverse relationship between precision and recall is known as the Precision-Recall Trade-off. This trade-off is shown in Figure \ref{fig:placeholder4} and Figure \ref{fig:placeholder5}, clearly. The model demonstrates strong performance across all four action categories, with "Move forward" and "Stop/slow down" achieving the highest average precision scores of 0.872 and 0.862, respectively. The performance varies significantly across different reason categories. "Follow traffic" achieves the highest average precision (AP=0.708), followed by "Road is clear" (AP=0.558) and "Traffic light is green" (AP=0.525), indicating the strong capability of the model in identifying these dominant reason categories. The majority of reason classes maintain precision above 0.6 at moderate recall levels, demonstrating the overall effectiveness of the model in multi-label reason classification. The curves exhibit the typical precision-recall trade-off, where higher recall values correspond to lower precision. It is generally impossible to maximize both metrics simultaneously, requiring practitioners to make strategic decisions based on application requirements.

\subsection{Statistical Complexity Analysis}
\begin{figure}
    \centering
    \includegraphics[width=1\linewidth]{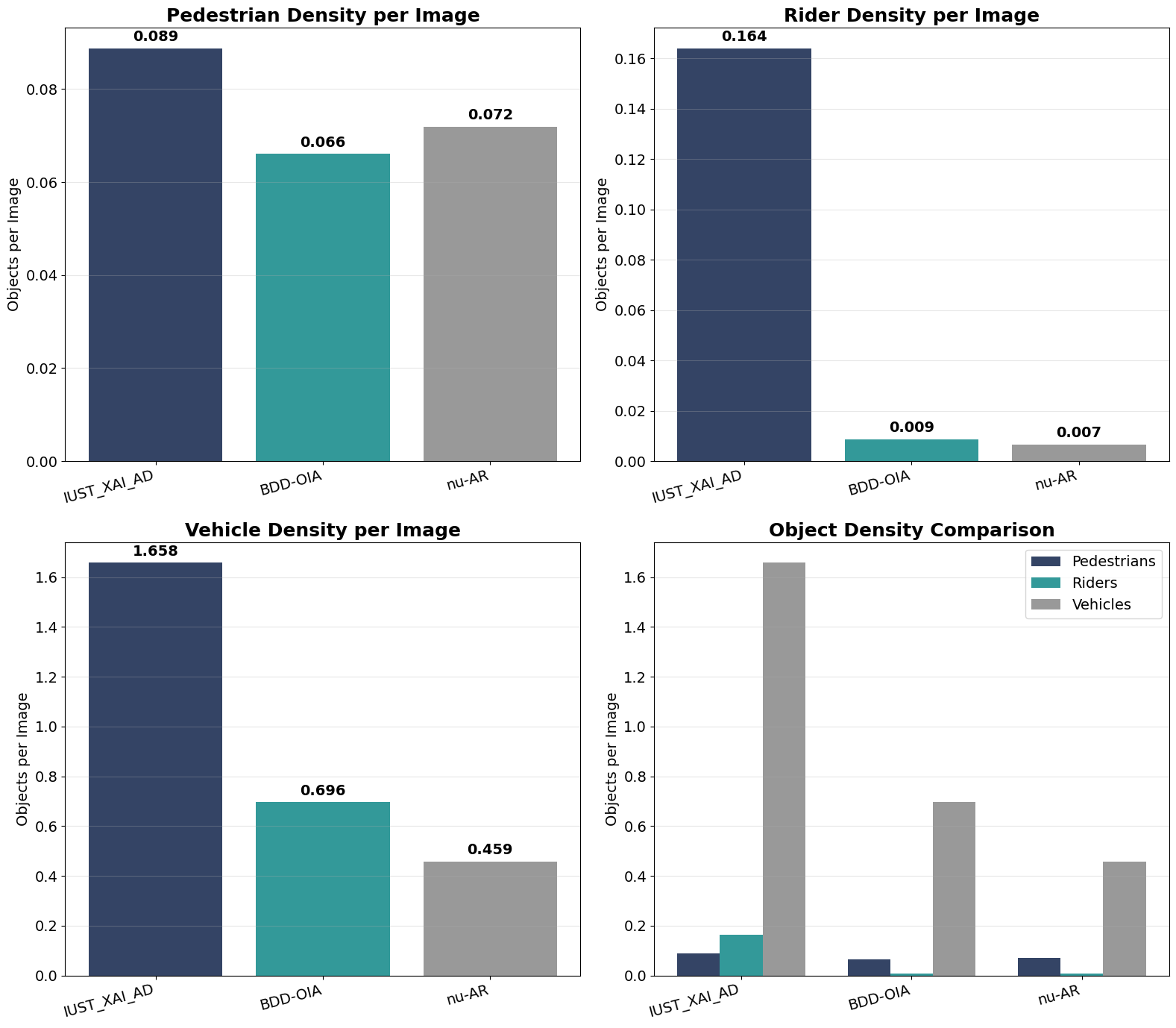}
    \caption{Detailed analysis of object type distributions across the three datasets, displaying Pedestrian Density, Rider Density, Vehicle Density per image, and a comparative overview of all object densities. IUST-XAI-AD shows the highest vehicle density (1.658) and rider density (0.164), while maintaining competitive pedestrian density (0.089) compared to other datasets.}
    \label{fig:placeholder6}
\end{figure}

\begin{figure}
    \centering
    \includegraphics[width=1\linewidth]{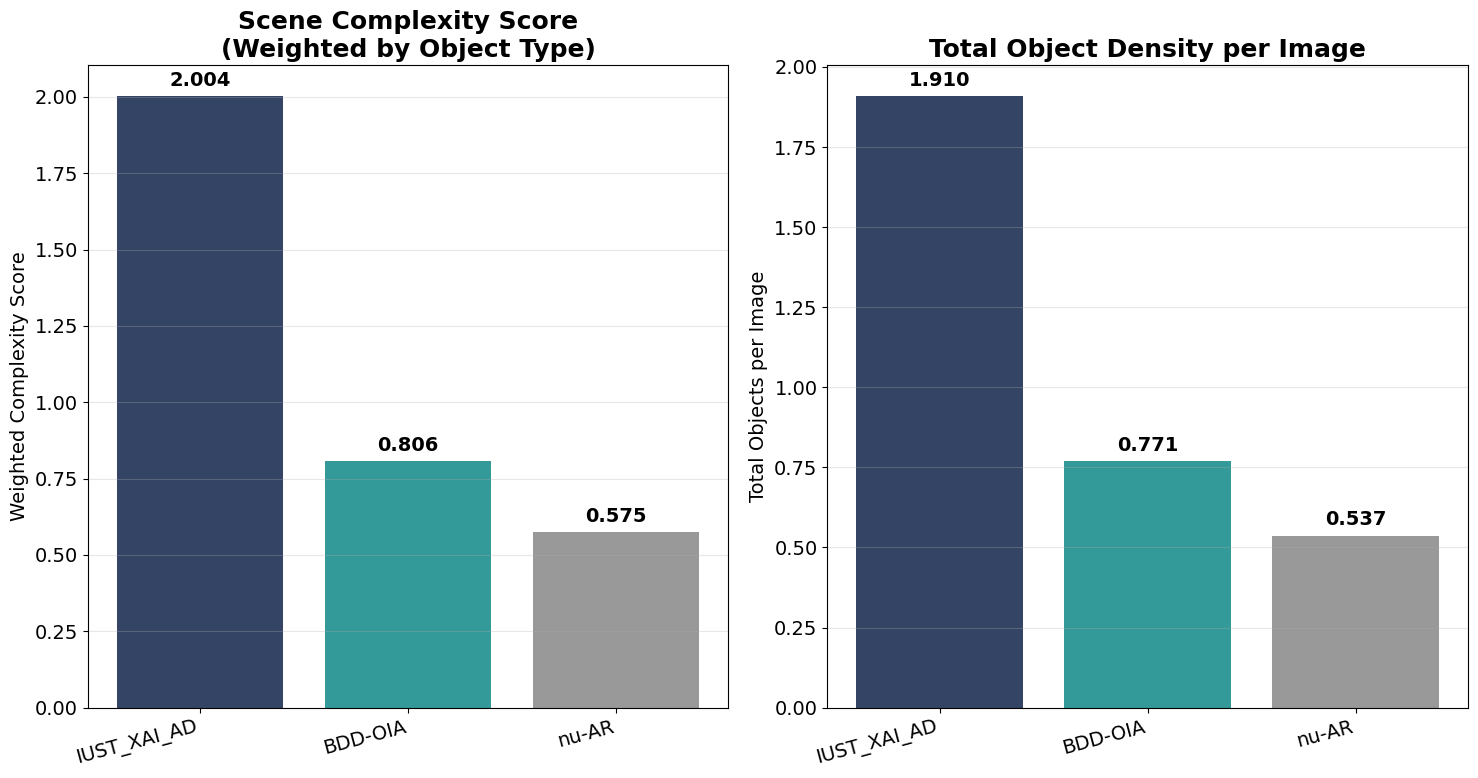}
    \caption{Comparison of dataset characteristics showing Scene Complexity Score (weighted by object type) and Total Object Density per Image across three datasets: IUST-XAI-AD, BDD-OIA, and nu-AR. IUST-XAI-AD demonstrates the highest complexity (2.004) and object density (1.910), followed by BDD-OIA and nu-AR, respectively.}
    \label{fig:placeholder7}
\end{figure}

Table \ref{tbl4} and Table \ref{tbl5} present a comprehensive statistical analysis comparing dataset complexity metrics. Additionally, Figure \ref{fig:placeholder6} and Figure \ref{fig:placeholder7} show the visual analysis. This analysis provides crucial insights into why IUST-XAI-AD presents greater challenges: IUST-XAI-AD demonstrates the highest scene complexity with a complexity score of 2.0038, significantly higher than BDD-OIA (0.8062) and nu-AR (0.5752). IUST-XAI-AD contains 1.66 vehicles per image, compared to 0.70 (BDD-OIA) and 0.46 (nu-AR), representing a 2.38 times and 3.61 times increase, respectively. Most notably, IUST-XAI-AD shows 0.164 riders per image compared to only 0.009 (BDD- OIA) and 0.007 (nu-AR), representing 19 times and 24 times higher density. This reflects the distinctive traffic patterns in Middle Eastern urban environments, where motorcycles and bicycles are more prevalent. IUST-XAI-AD exhibits 0.089 pedestrians per image, higher than both BDD-OIA (0.066) and nu-AR (0.072), indicating more pedestrian-rich scenarios.

\begin{table*}[width=2\linewidth,cols=6,pos=h]
\caption{Dataset Composition Showing the Distribution of Images and Object Counts across Different Categories (Pedestrians, Riders, Vehicles) for Three Datasets: BDD-OIA, nu-AR, and IUST-XAI-AD.}\label{tbl4}
\begin{tabular*}{\tblwidth}{@{} LLLLLL@{} }
\toprule
Dataset & Images & Pedestrians & Riders & Vehicles & Total Objects \\
\midrule
BDD-OIA \cite{xu2020explainable} & 4,572 & 302 & 40 & 3,181 & 3,523 \\
nu-AR \cite{feng2023nle} & 1,502 & 108 & 10 & 689 & 807 \\
IUST-XAI-AD & 958 & 85 & 157 & 1,588 & 1,830 \\
\bottomrule
\end{tabular*}
\end{table*}

\begin{table*}[width=2\linewidth,cols=6,pos=h]
\caption{Statistical Analysis of Scene Complexity across Datasets, Including Total Object Density, Vehicle Density, Rider Density, and Pedestrian Density. IUST-XAI-AD shows significantly higher complexity (2.0038) compared to BDD-OIA (0.8062) and nu-AR (0.5752), with particularly notable differences in rider density (19-24 times higher).}\label{tbl5}
\begin{tabular*}{\tblwidth}{@{} LLLLLL@{} }
\toprule
Dataset & Pedestrian Density & Rider Density & Vehicle Density & Total Density & Complexity \\
\midrule
BDD-OIA \cite{xu2020explainable} & 0.0661 & 0.0087 & 0.6958 & 0.7706 & 0.8062 \\
nu-AR \cite{feng2023nle} & 0.0719 & 0.0067 & 0.4587 & 0.5373 & 0.5752 \\
IUST-XAI-AD & 0.0887 & 0.1639 & 1.6576 & 1.9102 & 2.0038 \\
\bottomrule
\end{tabular*}
\end{table*}

The statistical analysis demonstrates that IUST-XAI-AD presents a more challenging evaluation benchmark. 2.5-3.5 times higher complexity scores indicate more demanding visual understanding requirements. Higher diversity in road users (vehicles, riders, pedestrians) requires more sophisticated reasoning capabilities. Unique traffic patterns provide insights into model generalization across different cultural contexts.

The analysis reveals significant cultural and regional differences in traffic patterns. The high rider density (19-24 times higher than Western datasets) reflects the mixed traffic conditions common in Middle Eastern cities, where motorcycles, bicycles, and cars share road space more intensively. The overall object density of 1.91 objects per image in IUST-XAI-AD is 2.5 times higher than BDD-OIA and 3.6 times higher than nu-AR, indicating denser urban environments typical of Middle Eastern cities. IUST-XAI-AD shows different action patterns with 56.1\% "Stop/slow down" actions compared to 46.0\% in BDD-OIA, suggesting more frequent stop-and-go traffic conditions.

This analysis positions IUST-XAI-AD as a valuable contribution to the autonomous driving research community, offering a more challenging and culturally diverse evaluation benchmark that complements existing Western-centric datasets. The higher complexity of the dataset and cultural specificity make it particularly suitable for testing model robustness and cross-cultural generalization capabilities.

\subsection{Performance Comparison Between Network Architectures on Newly and the Established Datasets}
Table \ref{tbl6} provides a comprehensive comparison between our previously model and the NLE-DM baseline across all three datasets. The results reveal consistent improvements over our previous model on IUST-XAI-AD across all metrics. Notably, our previous attention-based architecture demonstrates superior generalization and robustness compared to NLE-DM, outperforming it in 9 out of 12 metrics across the three datasets. This highlights the critical importance of evaluation on diverse and increasingly complex datasets, as not all models maintain consistent performance under varying conditions. The performance drop on IUST-XAI-AD compared to BDD-OIA suggests that distinct driving contexts present fundamentally different challenges that require specialized adaptation. The consistent performance degradation across all metrics for IUST-XAI-AD suggests inherent complexity in this dataset.

\begin{figure}
    \centering
    \includegraphics[width=1\linewidth]{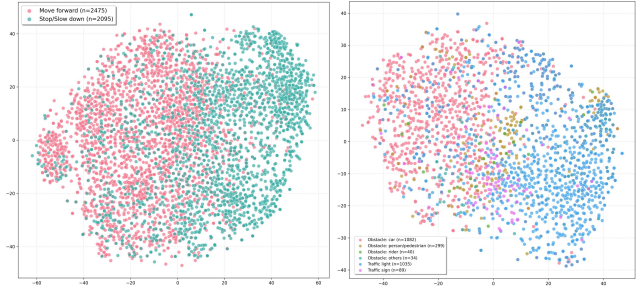}
    \caption{t-SNE visualization of learned feature representations. (Left) Feature embeddings colored by driving actions: Move forward (pink) vs. Stop/Slow down (teal). (Right) Same embeddings colored by obstacle types. The visualization reveals that the model learns feature representations that cluster by both action classes and environmental context, with clear separation between traffic infrastructure elements and dynamic obstacles.}
    \label{fig:placeholder8}
\end{figure}

Figure \ref{fig:placeholder8} shows two-dimensional t-SNE embedding of features colored by driving action classes: "Move forward" (pink, n=2475) and "Stop/Slow down" (teal, n=2095). The visualization shows clear clustering patterns with some overlap between classes, indicating that the model learns discriminative features while capturing the continuous nature of driving decisions. The t-SNE embedding for the reasons related to "stop/slow down" decision: car (pink, n=1082), person/pedestrian (orange, n=299), rider (green, n=40), others (yellow, n=34), traffic light (blue, n=1035), and traffic sign (purple, n=89). The feature space demonstrates that the model organizes representations based on both action semantics (left) and environmental context (right), with traffic infrastructure (lights and signs) forming distinct clusters while dynamic obstacles (cars, pedestrians) show more distributed patterns reflecting their varied behavioral contexts.

\begin{table*}[width=2\linewidth,cols=9,pos=h]
\caption{Performance Comparison Between Network Architectures. Comparison of F1-scores for attention-based architecture and NLE-DM model across three datasets (BDD-OIA, nu-AR, IUST-XAI-AD). Results show consistent performance differences between models, with both architectures demonstrating reduced performance on the more challenging IUST-XAI-AD dataset. The best results are specified in \textbf{bold}.}
\label{tbl6}
\begin{tabular*}{\tblwidth}{@{} L *{8}{C} @{} }
\toprule
Networks & \multicolumn{4}{c}{Attention-based \cite{azad2024xai}} & \multicolumn{4}{c}{NLE-DM \cite{feng2023nle}} \\
\cmidrule(lr){2-5} \cmidrule(lr){6-9}
Test Set & \makecell{F1-action \\ -overall} & \makecell{F1-action \\ -mean} & \makecell{F1-reason \\ -overall} & \makecell{F1-reason \\ -mean} & \makecell{F1-action \\ -overall} & \makecell{F1-action \\ -mean} & \makecell{F1-reason \\ -overall} & \makecell{F1-reason \\ -mean} \\
\midrule
BDD-OIA \cite{xu2020explainable} & 71.35 & 69.59 & \textbf{54.06} & \textbf{36.65} & \textbf{73.3} & \textbf{72.3} & 51.7 & 31.2 \\
nu-AR \cite{feng2023nle} & 69.67 & \textbf{70.24} & \textbf{59.38} & \textbf{35.77} & \textbf{73.3} & 68.8 & 49.9 & 30.8 \\
IUST-XAI-AD & \textbf{59.12} & \textbf{46.73} & \textbf{40.90} & \textbf{25.20} & 54.15 & 46.51 & 38.76 & 22.87 \\
\bottomrule
\end{tabular*}
\end{table*}

\section{Conclusions and Future Work}
\label{sec:section6}
This paper presents significant contributions to the field of explainable artificial intelligence for autonomous driving: a systematic confidence threshold sensitivity analysis, the introduction of the IUST-XAI-AD dataset, and a comprehensive validation of the dataset and the proposed threshold sensitivity analysis. Our systematic approach to threshold sensitivity analysis demonstrates that the conventional fixed threshold of 0.5 is suboptimal for multi-task scenarios. Our work demonstrates that threshold selection in multi-task autonomous driving should be viewed as a design choice rather than a pure optimization problem. The existence of a robust threshold range [0.3, 0.5] allows practitioners to balance multiple objectives (precision, recall, explainability, safety) based on deployment context, rather than optimizing a single predefined metric. The introduction of our culturally diverse dataset, IUST-XAI-AD, reveals important insights about cross-cultural driving behavior patterns and model generalization. The dataset provides a valuable benchmark for evaluating explainable computer vision in different cultural contexts. Comprehensive evaluation across multiple datasets and cultural contexts is essential for assessing the true robustness of autonomous driving systems.

In future work, we will investigate fine-tuning and threshold sensitivity analysis specifically for adverse weather scenarios (rain, fog, snow, low-light conditions). This is particularly critical for safety-critical decisions where environmental factors substantially impact sensor reliability and perception quality. While IUST-XAI-AD demonstrates the value of cultural diversity, comprehensive evaluation across multiple geographically and culturally diverse datasets would provide deeper insights into true model generalization capabilities. This would help identify specific cultural or infrastructural factors that impact model performance and guide the development of truly global autonomous driving systems.

\vspace{20pt}
\noindent \textbf{Data availability statement}
Three datasets were used in this study. The BDD-OIA dataset is available at \url{https://twizwei.github.io/bddoia_project/}. The nu-AR dataset is available at \url{https://github.com/lab-sun/NLE-DM}. The IUST-XAI-AD dataset and code are available on request.

\vspace{10pt}
\noindent \textbf{Declaration of competing interest}
The authors declare that they have no known competing financial interests or personal relationships that could have appeared to influence the work reported in this paper.

\vspace{10pt}
\noindent \textbf{Funding}
This research did not receive any specific grant from funding agencies in the public, commercial, or not-for-profit sectors.


\bibliographystyle{model1-num-names}

\bibliography{cas-refs}


\end{document}